\documentclass[11pt,a4paper]{article}
\usepackage[hyperref]{acl2019}
\usepackage{times}
\usepackage{latexsym}
\usepackage{amsmath}

\usepackage{hyperref}
\usepackage{url}

\usepackage{xcolor}
\usepackage{subcaption}

\usepackage{array, caption, tabularx, makecell}%
\usepackage{color}
\usepackage{colortbl}
\usepackage{graphicx}
\usepackage{booktabs}

\usepackage{multirow}

\aclfinalcopy 

\title{Maximizing Stylistic Control and Semantic Accuracy in NLG: \\ Personality Variation and Discourse Contrast}

\author{Vrindavan Harrison, Lena Reed, Shereen Oraby, Marilyn Walker \\
   Natural Language and Dialogue Systems Lab \\
   University of California Santa Cruz \\
  Santa Cruz, CA, US  \\
  {\tt \{vharriso, lireed, soraby, mawalker\}@ucsc.edu} \\
  }

\date{}

\begin{document}
\maketitle
\begin{abstract}
Neural generation methods for task-oriented dialogue typically
generate from a meaning representation that is populated using a
database of domain information, such as a table of data describing a
restaurant.  While earlier work focused solely on the semantic
fidelity of outputs, recent work has started to explore methods for
controlling the style of the generated text while simultaneously
achieving semantic accuracy.  Here we experiment with two stylistic
benchmark tasks, generating language that exhibits variation in
personality, and generating discourse contrast.  We report a huge
performance improvement in both stylistic control and semantic
accuracy over the state of the art on both of these benchmarks. 
We
test several different models and show that putting stylistic
conditioning in the decoder and eliminating the semantic re-ranker used in
earlier models results in more than 15 points higher BLEU for
Personality, with a reduction of semantic error to near zero. We also
report an improvement from	.75 to .81 in controlling contrast and a
reduction in semantic	error from 16\% to 2\%.

\end{abstract}

\section{Introduction}

\begin{table}[t]
    \centering
    \begin{small}
    \begin{tabular}{|p{7cm}|}  \toprule 
          \multicolumn{1}{|c|}{ \cellcolor[gray]{0.9}  \textbf{Meaning Representation} }  \\
            name{[}Browns Cambridge{]}, eatType{[}coffee shop{]}, 
            food{[}Italian{]}, 
            customerRating[average], 
            area[riverside], familyFriendly[yes], 
            near[Crowne Plaza Hotel]
            \\ \hline \hline
         \multicolumn{1}{|c|}{\cellcolor[gray]{0.9} \textbf{E2E Dataset} }\\
         \textit{Browns Cambridge is near Crowne Plaza Hotel. Browns Cambridge has a three star rating. Browns Cambridge is a family coffee shop.}\\ \hline
            \multicolumn{1}{|c|}{\cellcolor[gray]{0.9} \textbf{Personality: Conscientious}} \\ 
          \textit{Did you say Browns Cambridge? Well, i see, I think that it is a coffee shop, also it has a decent rating, and it is near Crowne Plaza Hotel kid friendly in riverside and an Italian place.}\\
          \hline
          \multicolumn{1}{|c|}{\cellcolor[gray]{0.9} \textbf{Personality: Disagreeable}} \\ 
          \textit{Come on, I am not sure. I mean Browns Cambridge is an Italian place, also it has a damn average rating. It is near Crowne Plaza Hotel.} \\ \hline
                    \hline

          \multicolumn{1}{|c|}{\cellcolor[gray]{0.9} \textbf{Personality: Unconscientious}} \\ 
          \textit{Oh God I don't know! Browns Cambridge is a coffee shop, also it is family friendly near Crowne Plaza Hotel, also it is an Italian place in riverside, also it has like, a decent rating. } \\ \hline
    \end{tabular}
\vspace{-.1in}
\caption{Sample meaning representation with a realization from the E2E Challenge Dataset and
two stylistic personality realizations.}
\label{table:mr-ref-example}
    \end{small}
    \vspace{-.2in}
\end{table}

Neural encoder-decoder models were originally developed for machine
translation \cite{sutskever2014sequence, Bahdanau_Cho_Bengio_2014},
but they have also been shown to be successful in related natural
language generation (NLG) tasks such as realizing dialogue system
utterances from meaning representations (MRs) as shown for the
restaurant domain in Table~\ref{table:mr-ref-example}
\cite{Dusek2016}.  Recent work in neural NLG has shown that stylistic
control is an important problem in its own right: it is needed to
address a well-known limitation of such models, namely that they
reduce the stylistic variation seen in the input, and thus produce
outputs that tend to be dull and repetitive \cite{li2016persona}.
Here we compare different methods for 
directly controlling stylistic variation when generating
from MRs, while simultaneously achieving high semantic accuracy.

Tables~\ref{table:mr-ref-example} and~\ref{table:mr-contrast-example}
illustrate the two stylistic benchmark datasets that form the basis of
our experimental setup.  Table~\ref{table:mr-ref-example} shows an
example MR with three surface realizations: the E2E realization does
not target a particular personality, while the other two examples vary
stylistically according to linguistic profiles of personality type
\cite{PennebakerKing99,Furnham90,MairesseWalker11}.
Table~\ref{table:mr-contrast-example} shows an example MR with two
surface realizations that vary stylistically according to whether the
discourse contrast relation is used
\cite{NakatsuWhite06,Howcroftetal13}.  Both of these benchmarks
provide parallel data that supports experiments that hold constant the
underlying meaning of an utterance, while varying the style of the
output text. In contrast, other tasks that have been used to explore
methods for stylistic control such as machine translation or
summarization (known as text-to-text generation tasks) do not allow
for such a clean separation of meaning from style because the inputs
are themselves surface forms.

\begin{table}[t]
    \centering
    \begin{small}
    \begin{tabular}{|p{7cm}|} \toprule 
          \multicolumn{1}{|c|}{\cellcolor[gray]{0.9} \textbf{Meaning Representation}} \\
            name{[}Brown's Cambridge{]}, 
            food{[}Italian{]}, 
            customerRating[3 out of 5], 
            familyFriendly[no], 
            price[moderate]
            \\
          \hline \hline
          \multicolumn{1}{|c|}{\cellcolor[gray]{0.9} \textbf{With Contrast Relation}} \\
          \textit{Browns Cambridge is an Italian restaurant with average customer reviews and \textbf{reasonable prices,  but it is not child-friendly.}}\\
          \hline
          \multicolumn{1}{|c|}{\cellcolor[gray]{0.9} \textbf{Without Contrast Relation}} \\ 
          \textit{Browns Cambridge serves Italian food in moderate price range. It is not kid friendly and the customer rating is 3 out of 5. } \\ \hline
    \end{tabular}
\vspace{-.1in}
\caption{A sample meaning representation with contrastive and non-contrastive surface realizations.}
\label{table:mr-contrast-example}
    \end{small}
\end{table}

We describe three methods of incorporating stylistic information as
\textit{side constraints} into an RNN encoder-decoder model, and test
each method on both the personality and contrast stylistic benchmarks.
We perform a detailed comparative analysis of the strengths and
weaknesses of each method.  We measure both semantic fidelity and
stylistic accuracy and quantify the tradeoffs between them.  We show
that putting stylistic conditioning in the decoder, instead of in the encoder as in previous work, and eliminating the
semantic re-ranker used in earlier models results in more than 15
points higher BLEU for Personality, with a reduction of semantic error
to near zero. We also report an improvement from .75 to .81 in
controlling contrast and a reduction in semantic error from 16\% to
2\%.  To the best of our knowledge, no prior work has conducted a
systematic comparison of these methods using such robust criteria
specifically geared towards controllable stylistic variation.  We
delay a detailed review of prior work to
Section~\ref{sec:related-work} when we can compare it to our own.

\section{Models and Variants}
\label{sec:model}
In the recent E2E NLG Challenge shared task, models were tasked with
generating surface forms from structured meaning
representations\cite{Dušek_Novikova_Rieser_2019}. The top performing
models were all RNN encoder-decoder systems. 
Our model also follows a standard RNN Encoder--Decoder model
\cite{sutskever2014sequence,Bahdanau_Cho_Bengio_2014} that maps a
source sequence (the input MR) to a target sequence.

\subsection{Model}
Our model represents an MR as a sequence $x = (x_1, x_2, \ldots x_n)$ of slot-value pairs. The generator is tasked with generating a surface realization which is represented as a sequence $y$ of tokens $y_1, y_2, \ldots y_m$. The generation system models the conditional probability $p(y|x)$ of generating the surface realization $y$ from some meaning representation $x$. Thus, by predicting one word at a time, the conditional probability can be decomposed into the conditional probability of the next token in the output sequence:
\begin{equation}
    p(y|x) = \prod_{t = 1}^{m} p(y_t| y_1, y_2, \ldots y_{t-1}, x) \; .
\end{equation}

We are interested in exercising greater control over the characteristics of the output sequence by incorporating \textit{side constraints} into the model \cite{Sennrich_Haddow_Birch_2016}. The side constraints $\textbf{c}$ act as an additional condition when predicting each token in the sequence. In this case, the conditional probability of the next token in the output sequence is given by:
\begin{equation}
    p(y|x, \textbf{c}) = \prod_{t = 1}^{m} p(y_t| y_1, y_2, \ldots y_{t-1}, x, \textbf{c})  \; .
\end{equation}
In Section \ref{sec:side-constraints} we describe three methods 
of computing  $p(y|x, \textbf{c})$ .


\paragraph{Encoder.}
The model reads in an MR as a sequence of slot-value pairs. Separate vocabularies for slot-types and slot values are calculated in a pre-processing step. Each slot type and slot value are encoded as one-hot vectors which are accessed through a table look-up operation at run-time. Each slot-value pair is encoded by first concatenating the slot type encoding with the encoding of its specified value.
Then the slot-value pair is encoded with an RNN encoder. 
We use a multi-layer bidirectional LSTM \cite{hochreiter1997long} 
to encode the input sequence of MR slot-value pairs. The hidden 
state $\bar{h_i}$ is represented as the concatenation of the forward 
state $\overrightarrow{h_i}$ and backward state $\overleftarrow{h_i}$. 
Specifically, $\bar{h_i} = (\overrightarrow{h_i},\overleftarrow{h_i})$ .


\paragraph{Decoder.} The decoder is a uni-directional LSTM. 
Attention is implemented as in \cite{luong2015effective}.
We use a global attention where the attention scores between two 
vectors $a$ and $b$ are calculated as $a^{T} \textbf{W} \, b$, 
where $\textbf{W}$ is a model parameter learned during 
training. 

\begin{figure}[t!]
	\centering
    \includegraphics[width=\linewidth]{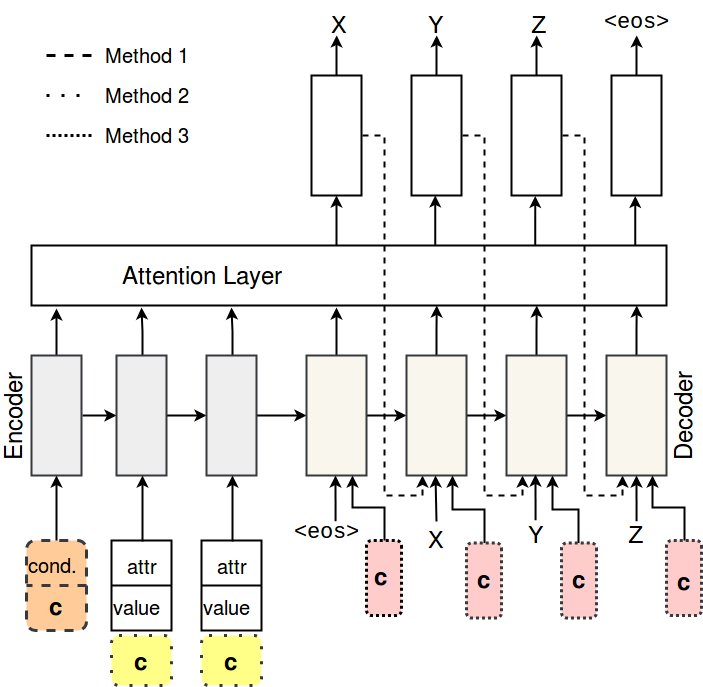}
    \caption{Attentional Encoder-Decoder architecture with each of the three side constraint implementations shown. The output sequence X, Y, Z is being generated from an MR represented as an input sequence of attribute value pairs.}
    \label{fig:nn-side-constraints}
\end{figure}

\subsection{Side Constraints}
\label{sec:side-constraints}
Recent work has begun to explore methods for stylistic control in
neural language generation, but there has been no systematic attempt
to contrast different methods on the same benchmark tasks and thereby
gain a deeper understanding of which methods work best and why.  Here,
we compare and contrast three alternative methods for implementing
side constraints in a standard encoder-decoder architecture.  The
first method involves adding slot-value pairs to the input MR, and the
second involves extending the slot-value encoding through a
concatenation operation. In the third method, side constraints are
incorporated into the model by modifying the decoder inputs.  The
three side constraint implementation methods are shown simultaneously
in Figure~\ref{fig:nn-side-constraints}. The orange area refers Method
1, the yellow areas corresponds to Method 2, and the red areas
corresponds to Method 3.

\label{sec:method-1}
\paragraph{Method 1: Token Supervision.}
This method provides the simplest way of encoding stylistic
information by  
inserting an additional
token that encodes the side constraint into the sequence of tokens
that constitute the MR \cite{Sennrich_Haddow_Birch_2016}. We add a new slot type representing
\texttt{side-constraint} to the vocabulary of slot-types, and
new entries for each of the possible side constraint values to the
vocabulary of slot values.

\label{sec:method-2}
\begin{figure}
	\centering
    \includegraphics[width=0.8\linewidth]{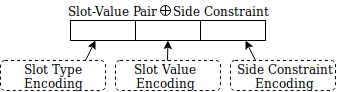}
    \caption{{\small Slot-value encoding extended with constraint.}}  
    \label{fig:slot-value-constraint}
\end{figure}

\paragraph{Method 2: Token Features.}
This method incorporates side constraints through use of a slot-value
pair feature. First we construct a vector representation $c$ that
contains the side constraint information. Normally the individual
slot-value pair encodings are built by concatenating the slot-type
with the slot-value as with Method 2. We modify
each slot-value pair encoding of the MR by extending it with $c$, as
seen in Figure~\ref{fig:slot-value-constraint}. 

\label{sec:method-3}
\paragraph{Method 3:~Decoder Conditioning.}
This method incorporates side constraint information into the
generation process by adding additional inputs to the LSTM decoder.
Traditionally, at the $t$-th time step a LSTM decoder takes two inputs. One
input is the previous ground truth token's embedding $w_{t-1}$, and
the other is a context vector $d_t$ which is an attention-weighted
average of the encoder hidden states.  A vector $c$ containing side
constraint information is provided to the decoder as a third
input. Thus at each time step the decoder's hidden state $\Tilde{h}_i$ is
calculated as
\begin{equation}
\Tilde{h}_i = \text{LSTM}([w_{t-1}; d_t; c]) \, .
\end{equation}
\vspace{-.1in}

%
%

%
%
\begin{table*}[ht]
\centering
\begin{footnotesize}
\begin{tabular}
{ |p{2.4cm}|p{12cm} |} \toprule
\bf Personality & \textbf{Realization}  \\ \midrule 
Meaning Representation & name[The Eagle], eatType[coffee shop], food[English], priceRange[cheap], customer rating[average], area[riverside], familyFriendly[yes], near[Burger King]
\\ \hline\hline
Agreeable & You want to know more about The Eagle? Yeah, ok it has an average rating, it is a coffee shop and it is an English restaurant in riverside, quite cheap near Burger King and family friendly.
\\ \hline
Disagreeable    & Oh god I mean, I thought everybody knew that The Eagle is cheap with an average rating, it's near Burger King, it is an English place, it is a coffee shop and The Eagle is in riverside, also it is family friendly.  
\\ \hline
Conscientious & I think that The Eagle is a coffee shop, it has an average rating and it is somewhat cheap in riverside and an English restaurant near Burger King. It is rather kid friendly.
\\ \hline
Unconscientious & Yeah, I don't know. Mmhm ... The Eagle is a coffee shop, The Eagle is cheap, it's kind of in riverside, it is an English place and The Eagle has an average rating. It is kind of near Burger King. \\ \hline
Extravert & The Eagle is a coffee shop, you know, it is an English place, family friendly in riverside and cheap near Burger King and The Eagle has an average rating friend!
\\ \hline
\end{tabular}
\vspace{-.1in}
\caption{Model outputs  for each personality style for a fixed Meaning Representation (MR). The model was trained using control Method 3.}
\label{table:personality-realization-examples}
\end{footnotesize}
\end{table*}

\section{Experiments: Varying Personality and Discourse Structure}
\label{sec:personality-experiment}
\label{sec:contrast-experiment}


We perform two sets of experiments using two stylistic benchmark
datasets: one for personality, and one for discourse structure, i.e., contrast.  In both cases, our aim is to generate stylized text from
meaning representations (MRs). In the personality experiments, the
generator's goal is to vary the personality style of the output and
accurately realize the MR.  The personality type is the side
constraint that conditions model outputs, and is
represented using a 1-hot encoding for the models that use side
constraint Methods 2 and 3. For the sake of comparison, we also train
a model that does not use conditioning ({\sc NoCon}).  In the
discourse contrast experiments, the generator's goal is to control
whether the output utterance uses the discourse contrast relation.
The side constraint is a simple boolean: contrast, or no contrast.
The model is tasked with learning 1) which category of items can
potentially be contrasted (e.g., \textit{price} and \textit{rating}
can appear in a contrast relation but \textit{name} can not), and 2)
which values are appropriate to contrast (i.e., items with polar
opposite values).

All models are 
implemented using PyTorch 
 and
OpenNMT-py\footnote{\url{github.com/OpenNMT/OpenNMT-py}}\cite{opennmt}. We use  
Dropout \cite{Srivastava_Hinton_Krizhevsky_Sutskever_Salakhutdinov_2014} 
of 0.1 between RNN layers. Model parameters are initialized using 
Glorot initialization \cite{Glorot_Bengio_2010} and are optimized using stochastic gradient descent 
with mini-batches of size 128. Beam search with three beams is used 
during inference. We implement multiple models for each experiment using the methods for 
stylistic control discussed in Section~\ref{sec:side-constraints}. We 
tune model hyper-parameters on a development dataset and select the model of 
lowest perplexity to evaluate on a test dataset. All models are trained 
using lower-cased and de-lexicalized reference texts. The sample model outputs we 
present have been re-capitalized and re-lexicalized using a simple rule based script. 
Further details on model implementation, hyper-parameter tuning, and data processing are provided as supplementary material. 

\subsection{Benchmark Datasets and Experiments}

\noindent\textbf{Personality Benchmark.} This dataset provides multiple reference outputs for each MR, where the
style of the output varies by personality type
\cite{Orabyetal18}.\footnote{\url{ nlds.soe.ucsc.edu/stylistic-variation-nlg}}
The styles belong to the
Big Five personality traits: agreeable, disagreeable, conscientious,
un-conscientious, and extrovert,  each with a stylistically
distinct linguistic profile
\cite{MairesseWalker10,Furnham90}. Example
model outputs for each personality on a fixed MR are
in Table~\ref{table:personality-realization-examples}.

The dataset consists of 88,855 train examples and 1,390 test examples
that are evenly distributed across the five personality types. Each
example consists of a (MR, personality-label, reference-text)
tuple. The dataset was created using the MRs from the E2E Dataset
\cite{e2e_dataset_novikova_duvsek_rieser_2017} and reference texts
synthesized by PERSONAGE \cite{mairesse2010towards}, a statistical
language generator capable of generating utterances that vary in style
according to psycho-linguistic models of personality. The statistical
generator is configured using 36 binary parameters that target
particular linguistic constructions associated with different
personality types.  These are split into {\it aggregation operations}
that combine individual propositions into larger sentences,
and {\it pragmatic markers} which typically modify some expression
within a sentence, e.g. {\it tag questions} or {\it in-group markers}.
A subset of these are illustrated in Table~\ref{table:agg-prag}: see
\citet{Orabyetal18} for more detail.

\begin{table}[htb!]
\begin{footnotesize}
\begin{tabular}
{@{} p{1.2in}|p{1.65in} @{}}
\hline
{\bf Attribute} & {\bf Example} \\ \hline\hline
\multicolumn{2}{l}{ \cellcolor[gray]{0.9} {\sc Aggregation Operations}}     \\               
{\sc "With" cue} &  {\it X is in Y, with Z.}  \\
{\sc Conjunction} & {\it X is Y and it is Z. \& X is Y, it is Z.} \\
{\sc "Also" cue} &  {\it X has Y, also it has Z.} \\
\multicolumn{2}{l}{ \cellcolor[gray]{0.9} {\sc Pragmatic Markers}}     \\    
{\sc ack\_justification} & \it I see, well \\ 
{\sc ack\_yeah} & \it yeah\\ 
{\sc confirmation} & 
{\it let's see ....., did you say X?  } \\ 
{\sc down\_kind\_of} & \it kind of \\ 
{\sc down\_like} & \it like \\ 
{\sc exclaim} & \it ! \\
{\sc general softener} & \it sort of, somewhat, quite, rather \\ 
{\sc emphasizer} & \it really, basically, actually, just  \\ 
{\sc tag question} & \it alright?, you see? ok?  \\ 
\hline
\end{tabular}
\end{footnotesize}
\vspace{-.1in}
\centering \caption{\label{table:agg-prag} 
Example Aggregation and Pragmatic Operations}
\end{table}


We conduct experiments using two control configurations that differ in
the granularity of control that they provide. We call the first
configuration \textit{course-grained} control, and the model is
conditioned using a single constraint: the personality label. The
second configuration, called \textit{fine-grained} control, conditions
the model using the personality label and Personage's 36 binary
control parameters as illustrated by Table~\ref{table:agg-prag}, which
provide fine-grained information on the desired style of the output
text. The stylistic control parameters are not updated during
training.  When operating under fine-grained control, for side
constraint Methods 2 and 3, the 1-hot vector that encodes personality
are extended with dimensions for each of the 36 control
parameters. For Method 1 we insert 36 tokens, one for each parameter,
to the beginning of each input sequence, in addition to the single
token that represents personality label.

\noindent\textbf{Contrast Benchmark.} This dataset provides reference outputs for 1000 MRs, where
the style of the output varies by whether or not it uses the discourse contrast relation.\footnote{\url{ nlds.soe.ucsc.edu/sentence-planning-NLG}}
Contrast training set examples are shown in Table~\ref{table:mr-contrast-example}.

The contrast dataset is based on
15,000 examples from the E2E generation challenge, which consists of
2,919 contrastive examples and 12,079 examples without contrast.\footnote{\url{www.macs.hw.ac.uk/InteractionLab/E2E/}} We
split the dataset into train and development subsets using a 90/10
split ratio. The test data is composed of a set of 500 MRs that
contain attributes that can be contrasted, whose reference outputs use
discourse-contrast
\cite{Reed_Oraby_Walker_2018}.
The test set also contains a set of 500 MRs that were selected from
the E2E development set that do not use discourse-contrast.  We crowd-sourced human-generated references for the
contrastive test set, and used the references from
the E2E dataset for the noncontrastive test set.\footnote{We will
  make our test and partitions of training data available to the
  research community if this paper is accepted.}

\subsection{Results} 

For both types of stylistic variation, we evaluate model outputs using
automatic metrics targeting semantic quality, diversity of the
outputs, and the type of stylistic variation the
model is attempting to achieve. We also conduct two human evaluations. In the tables and discussion that
follow, we refer to the models that employ each of the side constraint
methods, e.g., Methods 1, 2, and 3, described in
Section~\ref{sec:side-constraints}, using the monikers M\{1,2,3\}. The
model denoted NoCon refers to a model that uses no side constraint
information. Sample model outputs from the personality experiments are
shown in Table~\ref{table:personality-realization-examples}. The outputs are from the M3 model when operating under the fine grained control setting. 
Outputs from model M2 of the contrast experiment are shown in
Table~\ref{table:contrast-model-outputs}. 


\subsubsection{Semantic Quality} 
\label{sec:sem-quality}

%
%
\begin{table}[ht]
\centering
\begin{small}
\begin{tabular}{llllll}
\toprule
Model 			& BLEU  &  SER & H  & AGG & PRAG \\ 
\hline
\multicolumn{6}{c}{ \cellcolor[gray]{0.9} \citet{Orabyetal18}} \\
NoCon & 27.74 &  - &  7.87 & .56 &  .08\\
\textit{coarse} & 34.64 & - & 8.47 & .64 & .48\\
\textit{fine} & 37.66 &  - & 8.58 & .71 & .55 \\ 
\hline
\hline
\multicolumn{6}{c}{ \cellcolor[gray]{0.9} This Work} \\
Train & - & - & 9.34 & - & -  \\ 
NoCon & 38.45    & \textbf{0 } & 7.70  & .44 & .14 \\
\hline
\multicolumn{6}{c}{ \cellcolor[gray]{0.9} \textit{coarse control}} \\
M1  & 49.04    & \underline{0.000}  & 8.49 & .57 & .51 \\
M2  & 48.10    & 0.002  &  \underline{8.52} &.62  & .50 \\
M3  & \underline{49.06}     & 0.009  & 8.50  & .60 & .50  \\
\hline
\multicolumn{6}{c}{ \cellcolor[gray]{0.9} \textit{fine control}} \\
M1 & 55.30 &  \underline{0.004}  & 8.77 & .82 & .94 \\
M2 & 52.29 &  0.103  & \underline{\textbf{8.80}} & .84 & .93  \\ 
M3 & \textbf{55.98} &  0.014  & 8.74 & .84 & .93 \\
\bottomrule
\end{tabular}
\vspace{-.1in}
\caption{Automatic evaluation on Personality test set. \textit{course} and \textit{fine} refer to the specificity of the control configuration. \label{table:results-personality}}

\end{small}
\end{table}

First, we measure general similarity between model outputs and gold
standard reference texts using BLEU, calculated with the same
evaluation script\footnote{\url{github.com/tuetschek/e2e-metrics}} as
\citet{Orabyetal18}. For the personality experiment, the scores for each conditioning
method and each control granularity are shown in
Table~\ref{table:results-personality}, along with the results reported by \citet{Orabyetal18}. For the contrast experiment, the scores for
each conditioning method are shown in
Table~\ref{table:results-contrast}, where we refer to the model and results of
\citet{Reed_Oraby_Walker_2018} as \textit{M-Reed}. \citet{Reed_Oraby_Walker_2018} do
not report BLEU or Entropy (H) measures.

We first discuss the baselines from previous work on the same benchmarks.
Interestingly, for Personality, our {\sc NoCon} model gets a huge performance improvement 
of more than 11 points in BLEU (27.74 $\rightarrow$ 38.45) over results reported by \citet{oraby2018neural}. We note that while the underlying architecture behind our experiments 
is similar to the baseline described by \citet{oraby2018neural}, we experiment 
with different parameters and attention mechanisms.
\citet{Reed_Oraby_Walker_2018} and \citet{Orabyetal18} also use 
an LSTM encoder-decoder model with attention,
but they both implement their models using the  TGen\footnote{\url{github.com/UFAL-DSG/tgen}}\cite{duvsek2016context} framework 
with its default model architecture. 
TGen uses an early version of TensorFlow with different initialization 
methods, and dropout implementation. Moreover, we use a different 
one-hot encoding of slots and their values,
and we implement attention as in \citet{luong2015effective}, whereas 
TGen uses \citet{bahdanau2014neural} attention by default. 
Side constraints are incorporated into the
TGen models in two ways: 1) using a new dialogue act type
to indicate the side constraints, and 2) a feed-forward layer processes 
the  constraints and, during decoding, attention
is computed over the encoder hidden states and the hidden
state produced by the feed-forward layer. The TGen
system uses beam-search and an additional output re-ranking module. 

We now compare the performance of our own model results in Table~\ref{table:results-personality}.
As would be expected, NoCon has the lowest performance overall of all models, with a BLEU of 38.45.
With both coarse control and fine-grained control, M3 and M2 are
the highest and lowest performers, respectively.
For the contrast experiment, M2 and M3 have very similar values for
all rows of Table~\ref{table:results-contrast}. M2 has the highest
BLEU score of 17.32 and M3 has 17.09.  M1 is consistently outperformed
by both M2 and M3.  All side constraint models outperform NoCon. We
note that the contrast task achieves much lower scores on BLEU. This maybe due to the relatively small number of
contrast examples in the training set, but it is also possible that
this indicates the large variety of ways that contrast can be
expressed, rather than poor model performance. We show in a human
evaluation in Section~\ref{style-quality} that the contrast examples
are fluent and stylistically interesting.

A comparison of our results versus those
reported by \citet{Orabyetal18} are also shown in
Table~\ref{table:results-personality}. 
Note that our model has an over
14 point margin of improvement in BLEU score when using coarse control
and a more than 18 point improvement when using fine-grained control. Our models can clearly use the conditioning information more effectively than earlier work.

\begin{table}[ht!]
\begin{small}
\begin{center}
\begin{tabular}{cccc} \toprule
Model & BLEU   & SER & H  \\
\hline
Train   & -       &       & 10.68 \\ \hline
\multicolumn{4}{c}{\cellcolor[gray]{0.9} Contrast Data} \\
M-Reed  & -       & .16   &  -   \\
NoCon   & 15.80 & \textbf{.053}  & \textbf{8.09}  \\
M1      & 16.58 & .055  & 8.08   \\
M2      & \textbf{17.32} &  .058  & 8.03  \\
M3      & 17.09 &  .058  & 7.93  \\ \hline
\multicolumn{4}{c}{\cellcolor[gray]{0.9} Non Contrast Data} \\
NoCon   & 26.58 & .025  & 7.67  \\
M1      & 26.58 & .023 & 7.56  \\ 
M2      & 26.35 & .017 & 7.68  \\
M3      & 26.04 & .035 & 7.40  \\ \hline
\end{tabular}
\vspace{-.1in}
\end{center}
\end{small}
\caption{Automatic evaluation on Contrast test set. \label{table:results-contrast}}

\end{table}

\noindent
\textbf{Slot Error Rate.}
While the n-gram overlap metrics are able to measure general similarity between gold references and model outputs, they often do not do a good job at measuring semantic accuracy. Slot error rate (SER)\cite{wen2015semantically,Reed_Oraby_Walker_2018} is a metric similar to word error rate that measures how close a given realization adheres to its MR. 
SER\footnote{A formal definition of SER is provided in the supplementary materials.} 
is calculated 
using the slot aligner released\footnote{\url{github.com/jjuraska/slug2slug}} by \citet{Juraskaetal18} to
count the number of attributes (slots) and their values that correctly (and incorrectly) occur in a given surface realization. Please refer to Supplementary Materials, Section \ref{sec:appendix-calc-ser} for the definition of SER.

We evaluate each model using SER with results in Tables~\ref{table:results-personality} and \ref{table:results-contrast}. We first note that all the SERs for both tasks are extremely low
and that only M2 under fine control performs worse with an SER of .10. 
The models are clearly learning to realize the intended MRs. 
M1 has the best SER scores in all experiment conditions. 
In the contrast experiment, M2 and M3 are practically equivalent. 

\begin{table}[h!t]
\centering
\begin{small}
\begin{tabular}{lll} \toprule
Model   & Acc & Contrast Attempts \\ \hline
 M-Reed & .75 & 422  \\  \hline
M1      & .74 & 437 \\
M2      & .79 & 485 \\
M3      & .81 & 474 \\ \hline
\end{tabular}
\vspace{-.1in}
\end{small}
\caption{Contrast accuracy out of 500 examples.} \label{table:accuracy-results-contrast}
\end{table}

\begin{table*}[th]
   \centering
    \begin{small}
    \begin{tabular}{p{7cm}| p{8cm}} \toprule 
   \bf Meaning Representation & \textbf{Realization}  \\ \midrule 
name[Aromi], eatType[restaurant], rating[low], familyFriendly[yes]
    & \textit{Aromi is a \textbf{family friendly restaurant but the customer rating is low}.}
    \\ \hline
name[Fitzbillies], cuisine[English], 
price[more than \$30], eatType[pub], familyFriendly[yes]
    & \textit{Fitzbillies is a pub that serves English food, 
   \textbf{is children friendly, but the price range is more than \$30}.}
    \\ \hline
name[Clowns], price[more than \$30], rating[high], familyFriendly[no], near[Clare Hall]
    & \textit{Clowns is near Clare Hall. It has a \textbf{high customer rating but is not child friendly}.}
\\ \hline
name[Cotto], cuisine[English], location[riverside], price[high], eatType[coffee shop], rating[5 out of 5], near[The Portland Arms]
 & \textit{Cotto is a English coffee shop near The Portland Arms in the riverside. It has \textbf{a high price range but a customer rating of 5 out of 5}.}    
    \\ \bottomrule
   \end{tabular}
\vspace{-.1in}
\caption{Sample outputs from model M2 with contrast relation in bold.}
\label{table:contrast-model-outputs}
\end{small}
\end{table*}

\subsubsection{Quality in Variation}
\label{style-quality}
In the previous section we tested the ability of the side
constraint models to produce semantically accurate outputs. In this
section we evaluate the extent to which the side constraint models
produce stylistically varied texts. We evaluate variation using two
measures: 1) Entropy, and 2) counts on  model
outputs for particular stylistic targets.

\noindent\textbf{Entropy.}  Our goal is NLG
models that produce stylistically rich, diverse outputs, but we expect that variation in the training
data will be averaged out during model training. We quantify
the amount of variation in the training set, and also in the output
references from the test set MRs using 
Entropy\footnote{A formal definition of our Entropy calculation is provided with the supplementary materials.}, $H$, where a larger
entropy value indicates a larger amount of linguistic variation
preserved in the test outputs.  

The results are shown in the $H$ column of
Tables~\ref{table:results-personality} and
\ref{table:results-contrast}.  For the personality experiment, the
training corpus has 9.34 entropy and none of the models are able to
match its variability. When using fine-grained control M2 does the
best with 8.52 but all side constraint models are within 0.03. When
using coarse control M2 has the highest entropy with 8.80. 
Our models with fine control outperform
\citet{Orabyetal18} in terms of entropy. 
For the contrast experiment,
NoCon has the highest entropy at 8.09, but 
the differences are small.

\noindent\textbf{Counts of Stylistic Constructions.}  Entropy measures
variation in the corpus as a whole, but we can also examine the
model's ability to vary its outputs in agreement with the stylistic
control parameters. 
Contrast accuracy measures the ratio of valid contrast realizations to
the number of contrasts attempted by the
model. We determine valid contrasts using the presence of polar 
opposite values in the MR and then inspecting realization of 
those values in the model output.


Table~\ref{table:accuracy-results-contrast} shows the
results. The row labeled M-Reed refers to the results reported by
\citet{Reed_Oraby_Walker_2018}. NoCon rarely attempts contrast because there is no way to motivate it to do so, and it therefore produces no contrast.
Contrast attempts are out of 500 and
M2 has the most at 485. In terms of contrast accuracy M3 is the best
with 81\%.

When comparing our model performance to M-Reed, models M\{1,2,3\} 
make more contrast attempts. M1 and M-Reed have similar
contrast accuracy with 74\% and 75\%, respectively. The higher
performance of our models is particularly impressive since the M-Reed
models see roughly 7k contrast examples during training, which is
twice the amount that our models see.

For personality, we examine each model's ability to vary its outputs
in agreement with the stylistic control parameters by measuring
correlations between model outputs and test reference texts in the use
of the aggregation operations and pragmatic 
markers, two types of linguistic constructions illustrated
in Table~\ref{table:agg-prag}, and associated with each personality
type. 
The results for these linguistic constructions over all
personality types are shown in the last two columns (Agg, Prag) of
Table~\ref{table:results-personality}.  
The supplementary material
provides details for each personality. Our results demonstrate a
very large increase in the correlation of these markers between model
outputs and reference texts compared to previous work, and also
further demonstates the benefits of fine-grained control, where we
achieve correlations  to the reference texts as high as .94 for
pragmatic markers and as high as .84 for aggregation operations.

\noindent{\bf Methods Comparison.}
The results in Tables~\ref{table:results-personality} and
\ref{table:accuracy-results-contrast} reveal a general trend where
model performance in terms of  BLEU and entropy increases as
more information is given to the model as side constraints. At the
same time, the slot error rates are somewhat higher,
indicating the difficulty of simultaneously achieving 
both high semantic and stylistic fidelity.  Our conclusion is 
that Method 3 performs the best at controlling text style, but only when it has access to a large training dataset, and Method 2 performs better  in
situations where training data is limited. 


\noindent{\bf Human evaluation.}
We perform human evaluation of the quality of outputs for the M3 model with a random sample of 50 surface realizations for each personality, and 50 each for contrast and non-contrast outputs for a total of 350 examples. Three annotators on Mechanical Turk rate each output for both interestingness and fluency (accounting for both grammaticality and naturalness) using a 1-5 Likert scale.   

Human evaluation results are shown in Table~\ref{table:human-eval-personality} for the personality experiment and Table~\ref{table:human-eval-contrast} for contrast. The tables show average annotator rating in each category. For the personality outputs, each personality has similar fluency ratings with Conscientious slightly higher. The model outputs for the contrast relation have higher average ratings for Fluency than the non-contrastive realizations.  For interestingness, we compare both the personality styles and the contrastive style to the basic style without contrast. The results show that non-contrast (3.07), the vanilla style, is judged as significantly less interesting than the personality styles (ranging from 3.39 to 3.51) or the use of discourse contrast (3.45) (p-values all less than .01).

\begin{table}[ht]
\centering
\begin{small}
\begin{tabular}{lllllll} \toprule
& Con. 
& Dis.
& Agr.
& Ext.
& Unc.
& avg \\ \hline
Fluent
& 3.77 
& 3.38
& 3.53
& 3.38
& 3.35
& 3.48 \\
Interest 
& 3.39 
& 3.40
& 3.51
& 3.46
& 3.45
& 3.44 \\
\bottomrule
\end{tabular}
\vspace{-.1in}
\end{small}
\caption{Human evaluation results for personality.} \label{table:human-eval-personality}
\vspace{-.2in}
\end{table}

\begin{table}[ht]
\centering
\begin{small}
\begin{tabular}{lll} \toprule
& Non-contrast
& Contrast  \\ \midrule
Fluent
& 4.21
& 4.38 \\
Interest
& 3.07
& 3.45 \\
\bottomrule
\end{tabular}
\vspace{-.1in}
\end{small}
\caption{Human evaluation results for discourse contrast.} \label{table:human-eval-contrast}
\end{table}

\section{Related Work}
\label{sec:related-work}


Stylistic control is important as a way to address a well-known
limitation of vanilla neural NLG models, namely that they reduce the
stylistic variation seen in the input, and thus produce outputs that
tend to be dull and repetitive \cite{li2016persona}.  The majority of
other work on stylistic control has been done in a text-to-text
setting where MRs and corpora with fixed meaning and varying style are
not available
\cite{Fan_Grangier_Auli_2017,Iyyer_Wieting_Gimpel_Zettlemoyer_2018,
  Wiseman_Shieber_Rush_2018, Ficler_Goldberg_2017}. Sometimes
variation is evaluated in terms of model performance in some other
task, such as machine translation or summarization.
\citet{Herzig17} also control personality in the context of
text-2-text generation in customer care dialogues.
\citet{Kikuchi_Neubig_Sasano_Takamura_Okumura_2016} control output
sequence length by adding a remaining-length encoding as extra input
to the decoder.  \citet{Sennrich_Haddow_Birch_2016} control linguistic
honorifics in the target language by adding a special social formality
token to the end of the source
text. \citet{hu_controlled_generation_17} control sentiment and tense
(past, present, future) in text2text generation of movie reviews.
\citet{Ficler_Goldberg_2017} describe a conditioned language model
that controls variation in the stylistic properties of generated movie
reviews.

Our work builds directly on the approach and benchmark datasets of
\citet{Reed_Oraby_Walker_2018} and \citet{Orabyetal18}. Here we
compare directly to the results of \citet{Orabyetal18}, who were the
first to show show that a sequence-to-sequence model can generate
utterances from MRs that manifest a personality type.
\citet{Reed_Oraby_Walker_2018} also develop a neural model for a
controllable sentence planning task and run an experiment similar to
our contrast experiment. Here, we experiment extensively with
different control methods and present large performance improvements
on both tasks.


\section{Conclusion}
\label{sec:conclusion}

We present three different models for stylistic control of an attentional encoder-decoder model that generates restaurant descriptions from structured semantic representations using two stylistic benchmark datasets: one for personality variation and the other for variation in discourse contrast. We  show that the best models can simultaneously  control the variation in style while maintaining semantic fidelity to a meaning representation. Our experiments suggest that overall, incorporating style information into the decoder performs best and we report a large  performance improvement on both benchmark tasks, over a large range of metrics specifically designed to measure semantic fidelity along with stylistic variation. A human evaluation shows that the outputs of the best models are judged as fluent and coherent and that the stylistically controlled outputs are rated significantly more interesting than more vanilla outputs.


\bibliography{mybib,nl,phd}

\bibliographystyle{acl_natbib}

\newpage

\appendix

\section{Supplementary Materials: Maximizing Stylistic Control and Semantic Accuracy in Dialogue Generation:
Conditional Decoding for Personality Variation and Discourse Contrast}
\label{sec:appendix}

\subsection{Calculating Slot Error Rate}
\label{sec:appendix-calc-ser}

Multiple methods of measuring SER have been proposed \cite{wen2015semantically,Reed_Oraby_Walker_2018}. In this work we use a method similar to the one described by \citet{Reed_Oraby_Walker_2018}. 
First, we define the following types of errors: substitutions (realizing an attribute with an incorrect value), deletions (failing to mention an attribute), repeats, and hallucinations (mentioning an attribute that does not appear in the MR). 

The SER score for a given (MR, text realization) pair is calculated by first calculating $S$, $D$, $R$, and $\Tilde{H}$, which are the amounts of substitutions, deletions, repeats, and hallucinations, respectively. The SER formula is then given as:
\begin{equation}
    \text{SER} = \frac{S + D + R + \Tilde{H}}{N}
\end{equation}
where $N$ is the number of slots in the MR. Note that using this method can result in SER values greater than one, since it is possible for there to be more errors than slots in the MR.

\subsection{Calculating Entropy}
\label{sec:appendix-calc-entropy}

To calculate Shannon Text Entropy $H$, we first construct the corpus vocabulary $V$ of all unigrams, bigrams, and trigrams. Then $H$ is given by the equation
\begin{equation}
H = - \sum_{a \in V} \frac{k_a}{N} \cdot \log_{2}(\frac{k_a}{N})
\end{equation}
where $N$ is the sum total of occurrences for all terms in $V$, and $k_a$ is the number of occurrences for the term $a$.

\subsection{Model Implementation Details}
\label{sec:appendix-model-implementation}

\noindent \textbf{Model Implementation.} All models are 
implemented using PyTorch
 and
OpenNMT-py\footnote{\url{github.com/OpenNMT/OpenNMT-py}} 
 \cite{opennmt}. We use  
Dropout \cite{Srivastava_Hinton_Krizhevsky_Sutskever_Salakhutdinov_2014} 
of 0.1 between RNN layers. Model parameters are initialized using 
Glorot initialization \cite{Glorot_Bengio_2010} and are optimized using stochastic gradient descent 
with mini-batches of size 128. Beam search with three beams is used 
during inference. We implement multiple models for each experiment using the methods for 
stylistic control discussed in Section~\ref{sec:side-constraints}. We 
tune model hyper-parameters on a development dataset and select the model of 
lowest perplexity to evaluate on a test dataset. All models are trained 
using lower-cased and de-lexicalized reference texts. The sample model outputs we 
present have been re-capitalized and re-lexicalized using a simple rule based script. 

\textbf{Hyper Parameter Tuning.} Hyper parameters are tuned using a grid search over the following parameter space:
\begin{itemize}
    \item RNN layers over the range [1, 2] 
    \item RNN size over the range [150, 200, 250, 300]
\end{itemize}
We tune the number RNN layers and RNN size by training a model for each combination of layers and RNN size (8 models). We use the model of lowest development dataset perplexity to evaluate on the test dataset.

This parameter tuning process is performed for each of the side constraint methods and style parameter configuration (fine control, coarse control). The resulting hyper parameter values are shown in Table~\ref{table:hyper-parameters} 

\begin{table}[ht]
\begin{center}
\begin{tabular}{lll}
\toprule
Model & RNN layers  & RNN size \\ 
\hline
NoCon & 2 & 150  \\
\hline
\multicolumn{3}{c}{ \cellcolor[gray]{0.9} coarse control} \\
M1 & 1 & 200  \\
M2 & 1 & 200  \\
M3 & 2 & 150  \\

\hline
\multicolumn{3}{c}{ \cellcolor[gray]{0.9} fine control} \\
M1 & 1 & 200 \\
M2 & 2 & 200 \\
M3 & 1 & 200 \\

\bottomrule
\end{tabular}
\caption{Model hyper-parameter values.}
\label{table:hyper-parameters}
\end{center}
\end{table}

\subsection{Data Processing}
\label{sec:appendix-data-processing}
The data is pre-processed using Stanford CoreNLP \cite{Manningetal14}. 

\subsection{Linguistic constructions: Pragmatic Markers and Aggregation Operations}

Psycholinguistic studies have shown these markers to be indicative of the language of people
with different personality traits \cite{PennebakerKing99,Furnham90}.
For example, the 
use of pragmatic markers has been shown to effect perceptions of personality traits such as politeness, friendliness, extraversion, and enthusiasm \cite{oberlander2004individual, levinson1987politeness,dewaele1999extraversion}. Using a method similar to  \citet{Orabyetal18}, we count the occurrences of pragmatic markers and aggregation operations in the model outputs. Then we average the counts within each personality category and calculate the Pearson correlation between the model output averages and the gold reference text averages. 

The Pearson correlation $r$ for pragmatic markers can be seen in Table~\ref{table:results-prag-correlations}. All values of $r$ are significant with $p$-values less than $0.01$. The model with no side constraints has $r \leq 0.17$ for all personalities except for conscientious with $r = 0.81$. This suggests that the un-constrained model picks one personality to optimize -- conscientious in this case. For both control granularities each of the side constraint models have similar performance. Table~\ref{table:results-prag-correlations} also shows the correlation results reported by \citet{Orabyetal18} where we observe a marked improvement in the pragmatic marker correlations of our models compared to theirs.

Pearson correlations for aggregation operations are shown in Table~\ref{table:results-agg-correlations}. Again, the test for correlation results in $p$-values less than $0.01$ for each personality type.   Here, the Token model of \citet{Orabyetal18} outperforms all three of our models when conditioning on only the personality label (coarse control).

%
%
\begin{table}[ht]
\centering
\begin{small}
\begin{tabular}{@{}lllllll@{}}
\toprule
Model & AGR & CON & DIS & EXT & UNC & avg \\
\hline
\multicolumn{7}{c}{\cellcolor[gray]{0.9} Oraby et al} \\
NoSup & 0.05 & 0.59 & -0.07 & -0.06 & -0.11 & .08 \\
Token & 0.35 & 0.66 & 0.31 & 0.57 & 0.53 &  .48 \\
Context & 0.28 & 0.67 & 0.40 & 0.76 & 0.63 & .55 \\
\hline
\multicolumn{7}{c}{\cellcolor[gray]{0.9} This Work - coarse control} \\
NoCon & .17 & .81 & -.08 & -.08 & -.11 & .14\\
M1 & .44 & .81 & .17 & .79 & .32 & .51\\
M2 & .44 & .81 & .17 & .83 & .27 & .50\\
M3 & .40 & .81 & .14 & .83 & .31 & .50\\
\hline
\multicolumn{7}{c}{\cellcolor[gray]{0.9} This Work - fine control} \\
M1 & .87 & .94 & .98 & .99 & .90 & .94 \\
M2 & .87 & .94 & .98 & .99 & .88 & .93 \\
M3 & .87 & .93 & .97 & .99 & .90 & .93 \\
\bottomrule
\end{tabular}
\end{small}
\caption{Correlations between test examples and model outputs for pragmatic markers.}
\label{table:results-prag-correlations}
\end{table}

%
%
\begin{table}[ht]
\centering
\begin{small}
\begin{tabular}{@{}lllllll@{}}
\toprule
Model & AGR & CON & DIS & EXT & UNC & avg \\
\hline
\multicolumn{7}{c}{\cellcolor[gray]{0.9} Oraby et al} \\
NoSup & 0.78 & 0.80 & 0.13 & 0.42 & 0.69 & .56 \\
Token & 0.74 & 0.74 & 0.57 & 0.56 & 0.60 & .64 \\
Context & 0.83 & 0.83 & 0.55 & 0.66 & 0.70 & .71 \\
\hline
\multicolumn{7}{c}{\cellcolor[gray]{0.9} This Work - coarse control} \\
NoCon & 0.70 & 0.73 & -0.19 & 0.35 & 0.60 & .44 \\
M1 & 0.67 & 0.70 & 0.58 & 0.56 & 0.36 & .57 \\
M2 & 0.61 & 0.70 & 0.58 & 0.60 & 0.60 & .62 \\
M3 & 0.64 & 0.68 & 0.58 & 0.59 & 0.49 & .60 \\
\hline
\multicolumn{7}{c}{\cellcolor[gray]{0.9} This Work - fine control} \\
M1 & 0.84 & 0.91 & 0.78 & 0.81 & 0.78 & .82 \\
M2 & 0.89 & 0.92 & 0.78 & 0.79 & 0.84 & .84 \\ 
M3 & 0.86 & 0.91 & 0.79 & 0.82 & 0.81 & .84\\ 
\bottomrule
\end{tabular}

\end{small}
\caption{Correlations between test examples and model outputs for aggregation operations.}
\label{table:results-agg-correlations}
\end{table}

\end{document}